\documentclass[conference]{IEEEtran}
\IEEEoverridecommandlockouts
\usepackage{cite}
\usepackage{amsmath,amssymb,amsfonts}
\usepackage{algorithmic}
\usepackage{graphicx}
\usepackage{textcomp}
\usepackage{xcolor}
\usepackage{float}
\usepackage{tikz}
\usepackage{pgfplots}
\usepackage{url}
\pgfplotsset{compat=1.18}

\def\BibTeX{{\rm B\kern-.05em{\sc i\kern-.025em b}\kern-.08em
    T\kern-.1667em\lower.7ex\hbox{E}\kern-.125emX}}
\begin{document}

\title{IVG-UAV: An Intelligent Voice-Guided UAV System for Autonomous Ripe Fruit Harvesting with Vision-Based Classification and Adaptive Path Planning\\
}

\author{%
  \IEEEauthorblockN{Dinh Trung Duong}%
  \IEEEauthorblockA{\textit{School of Computing}\\
  \textit{SUNY Binghamton University}\\
  Binghamton, United States\\
  dduong3@binghamton.edu}%
}
\maketitle

\begin{abstract}
In tropical regions, their agricultural sectors remain highly dependent on manual labor for fruit harvesting. On large-scale farms, this dependency often results in significant labor cost and logistic complexities. This project presents the development and simulation of a voice-controlled Unmanned Aerial Vehicle (UAV) system designed to automate harvesting tasks in extensive plantations. The proposed system integrates speech recognition using Whisper\cite{radford2023robust} and LLM, computer vision–based ripeness classification, and adaptive path planning within a unified framework. The entire system is modeled and validated in a Gazebo simulation environment, allowing performance evaluation under controlled agricultural scenarios.
\end{abstract}


\section{Introduction}
In tropical nations, their agricultural sectors remain highly dependent on manual labor for fruit harvesting. On large-scale farms, this dependency often results in significant labor cost and logistic complexities. With recent developments, robots have been introduced to reduce this burden as well as improve the productivity. A diversity of harvesting robots have been significantly investigated ranging from ground-based mobile manipulators to aerial platforms such as unmanned aerial vehicles (UAVs). These systems are mostly equipped with vision-assisted technologies, enabling fruit detection, ripeness classification, and localization through advanced computer vision and deep learning algorithms. However, the majority of existing robots depends on fully automated programming paradigms, where robots autonomously navigate and search for ripe fruits based solely on pre-defined algorithms and perception models. This leads to the lack of real-time human interaction and adaptability, particularly during the fruit-searching and decision-making stages. This functionality allows the robot to adapt its flight trajectory according to user instructions. Additionally, it enables the UAV to handle edge cases through human intervention. This project proposes the integration of voice-guided interaction into the robotic system. This human-in-the-loop approach allows adaptive decision-making during the harvesting process, which enhances operational flexibility. To address the existing challenges, this work introduces IVG-UAV, a unified voice-guided UAV harvesting pipeline that integrates speech recognition, LLM-based command interpretation, vision-based fruit detection using YOLOv8 and RGB-D sensing, and adaptive path planning for autonomous navigation in orchard environments.
\begin{figure}[H]
\centering
\includegraphics[width=\linewidth]{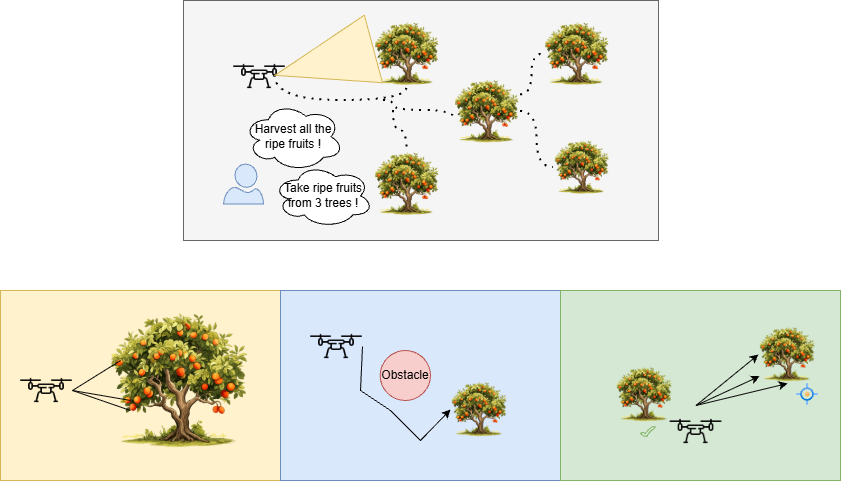}
\caption{UAV system will integrate voice command to automate the harvesting process. Using laser beam equipped with vision-based ripeness classification to recognize the satisfied objects and avoid obstacle during the harvesting process. Applying RRT* algorithm to find the shortest path aimed to the target}
\label{fig:uav-system}
\end{figure}
Figure 1 illustrates the overall functionality of the proposed voice-guided interactive harvesting system. The framework integrates aerial perception, obstacle-aware navigation, adaptive task planning, and human-in-the-loop voice interaction into a unified architecture for intelligent fruit harvesting. Initially, the UAV performs journey exploration of the orchard and scans fixed trees using vision system. Robot performs a thorough scan by applying row-following strategy, that ensures all the trees in the farm are scanned. Unlike fully autonomous harvesting systems, the proposed framework enables real-time human guidance through voice commands. The operator can issue high-level instructions such as “Harvest all the ripe fruits” or “Take ripe fruits from three trees.” These commands are processed by a speech recognition module and translated into structured task constraints that influence the target selection and planning modules. During visiting a tree, a drone system can use vision perception to classify the ripe fruits and can actually detach them from the tree. In a flight journey, UAV system can avoid obstacles to find a target tree and using RRT* \cite{karaman2011sampling} algorithm to find the shortest path that leads to the next target tree.

\section{Related works}

\subsection{Vision-assisted Avocado Harvesting UAV}
In order to harvest avocados, this paper developed a vision-assisted robotic harvesting system which utilizes an aerial robot with a bimanual manipulation mechanism. The device integrates a dual-arm UAV manipulator with a gripper arm and a fixer arm with computer vision for avocado detection and posture estimation. A motion planner uses the location and orientation of the avocados determined by the visual perception module to direct the robotic arms. The gripper arm rotates to effectively separate the avocado while the fixer arm holds the fruit stem in place during harvesting. The system can recognize and pick avocados on its own in outdoor settings, according to experimental data, demonstrating the promise of aerial robotic manipulation for agricultural automation.\cite{alaaudeen2024intelligent}
\subsection{Deep learning–based computer vision with robotic manipulation}
This work suggested an intelligent robotic harvesting system that combines robotic manipulation and deep learning-based computer vision for automated fruit picking. In order to increase harvesting efficiency, the system focuses on forecasting the best fruit gripping locations. In order to accomplish this, the authors created an image processing pipeline based on deep learning that recognizes apples in pictures and determines the proper gripping area for robotic picking. For fruit recognition, a lightweight one-step object detection network is used, allowing the system to instantly recognize fruits from visual input. The robotic manipulator is then guided by the detected fruit attributes to identify appropriate grabbing places and carry out the harvesting process.\cite{liu2024vision}

Both studies focus on automating fruit harvesting using computer vision–based perception and robotic manipulation techniques. In particular, they utilize visual sensing to detect fruits, estimate their position, and guide robotic systems to perform harvesting tasks. Although both strategies significantly rely on computer vision technology, their system architectures are different: the latter focuses on vision-based grasp prediction utilizing lightweight deep learning models for robotic manipulators, whereas the former incorporates aerial robotics with bi-manual manipulation. Despite these differences, both works share similar limitations. The systems lack natural human–robot interaction mechanisms, such as voice-guided instructions, which could allow users to intuitively guide or supervise the harvesting process. This limitation reduces system usability in practical agricultural environments where human workers may need to interact with robotic systems dynamically.

\section{Methodology}
Motivated by the limitations of existing systems, the author proposes a unified framework that combines voice-guided commands with existing robotic harvesting technologies. This approach introduces human-in-the-loop interaction, enabling users to provide intuitive voice instructions and improving the overall efficiency of robotic harvesting systems.
The solution contains two phases: exploration and mission execution. During the exploration phase, the UAV uses coverage-based navigation to look over the orchard and make a map of the area by finding the locations of the trees. In the next phase of the mission, the system uses the map it created to plan the best ways to harvest and detect the ripe fruits.
\subsection{Exploration Phase}
\begin{figure}[H]
\centering
\includegraphics[width=\linewidth]{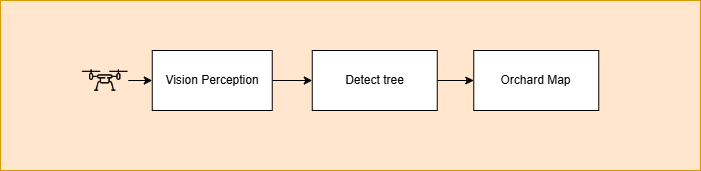}
\caption{Exploration and orchard mapping flow. The UAV performs exploration flight, the vision module detects trees, and the orchard map is constructed from the detected tree positions}

\label{fig:exploration}
\end{figure}
In the exploration phase, the UAV utilizes an RGB-D camera to observe the surrounding environment and detect trees within the orchard. 

\begin{figure}[H]
\centering
\includegraphics[width=\linewidth]{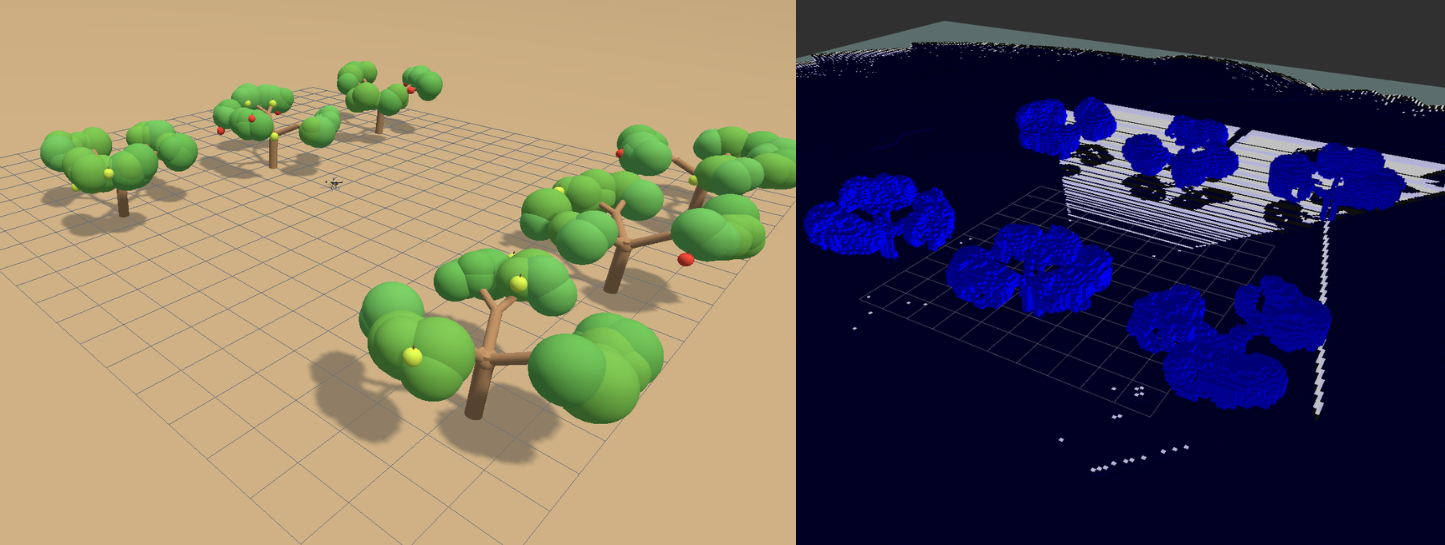}
\caption{Gazebo environment conversion into probabilistic voxel-based 3D map using Octomap \cite{hornung13auro}}

\label{fig:exploratio1n}
\end{figure}
The RGB-D sensor continuously captures both color and depth information, where the depth measurements are integrated into an OctoMap \cite{hornung13auro} framework to generate a probabilistic voxel-based 3D map of the orchard. Each voxel stores occupancy information, allowing the system to model tree trunks in a consistent spatial representation. As it moves around the orchard, the UAV gradually updates the OctoMap row by row and tree by tree. The produced voxel map allows for dependable spatial knowledge from a variety of perspectives while maintaining the environment's geometric integrity. The system determines each tree's spatial location in the global coordinate frame based on the occupied voxel clusters that correspond to tree structures. Each detected tree 's spatial position is represented by a 3D coordinate 
\[
T_i = (x_i, y_i, z_i),
\]
where \(x_i\), \(y_i\), and \(z_i\) denote the spatial coordinates of the \(i\)-th tree center within the orchard map.

\subsection{Mission execution phase}
\begin{figure}[H]
\centering
\includegraphics[width=\linewidth]{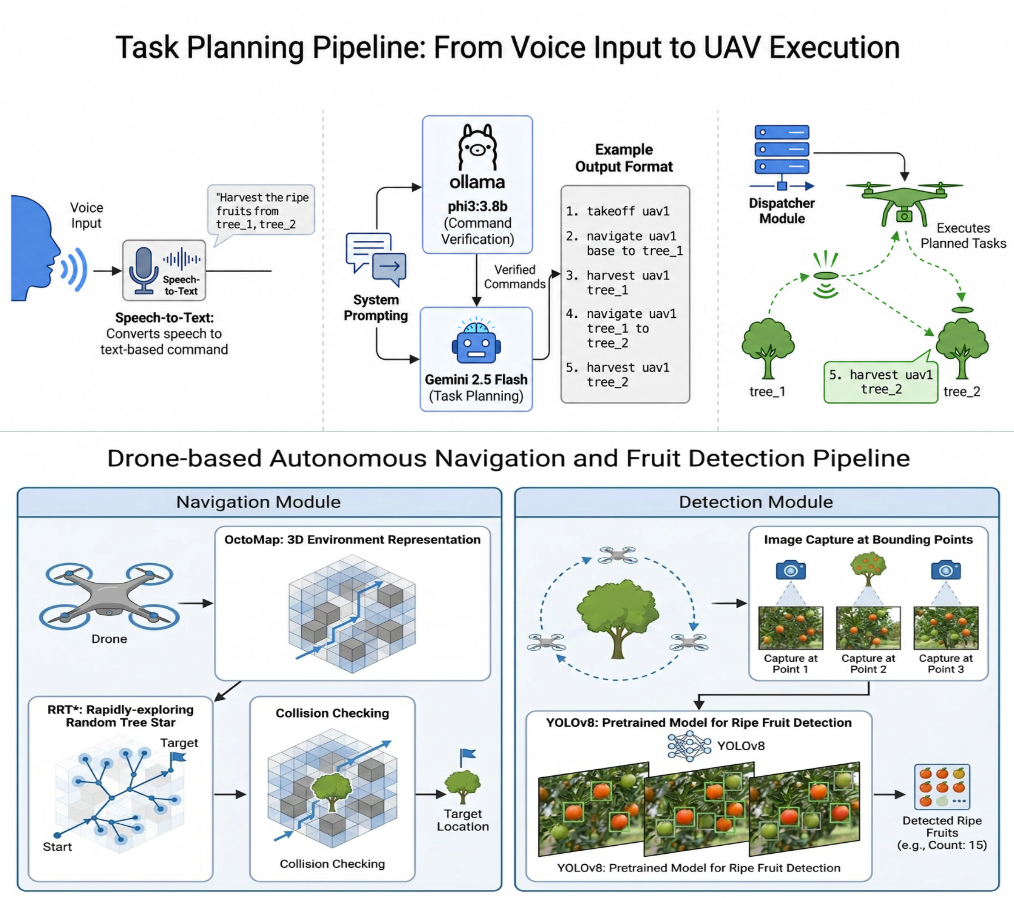}
\caption{The mission execution pipeline starts with voice commands, which are converted into mission instructions using Whisper and an LLM-based interpreter. The adaptive path planner generates navigation commands while handling obstacle avoidance. MAVSDK transmits control commands to the UAV, while the vision perception module continuously processes visual data during flight. The system runs on ROS2 within a Gazebo simulation environment.}

\label{fig:mission execution}
\end{figure}

After obtaining the orchard map from the exploration phase, the system enters the mission execution state. The execution pipeline is divided into three major components: (1) command generation and task planning, (2) navigation module, and (3) detection module.
\newline

\subsubsection{Command generation and task planning}
The command generation and task planning module converts user instructions into structured executable UAV actions. In general, the system receives voice commands from the human operator and utilizes the Whisper \cite{radford2023robust} speech recognition model to convert the spoken instructions into text-based commands. The generated textual input is subsequently processed by a multi-layer large language model (LLM) framework that integrates both Phi-3.3B \cite{abdin2024phi3technicalreporthighly} and Gemini 2.5 Flash models. By incorporating current mission context and domain-specific knowledge, the framework performs semantic verification, edge-case analysis, and hierarchical task planning to generate a structured UAV action plan for autonomous mission execution.

\begin{figure}[H]
\centering
\includegraphics[width=\linewidth]{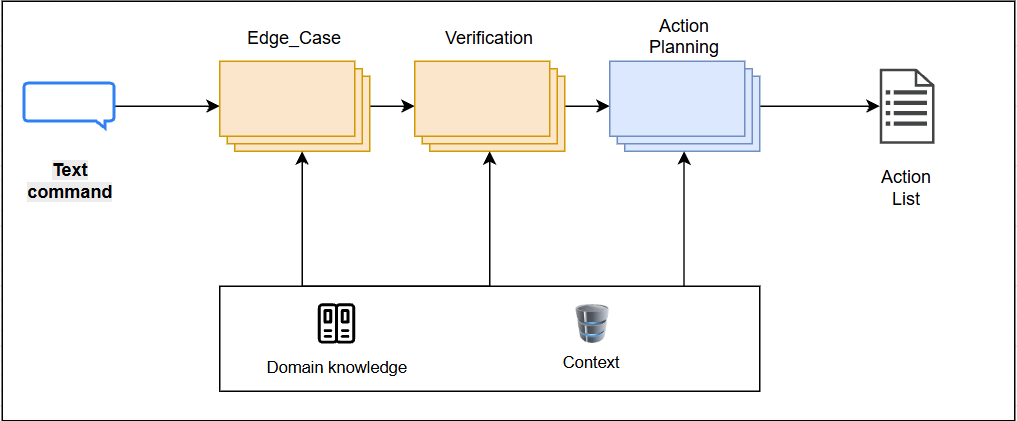}
\caption{Multi-architecture LLM layers}

\label{fig:llm}
\end{figure}

As illustrated in Figure \ref{fig:llm}, the Edge\_Case layer is implemented using the Phi-3.3B model, which utilizes current mission context and domain-specific knowledge to identify ambiguous instructions, invalid targets, and handle unclear situations such as commands containing relative descriptions including “furthest”, “nearest” or “all”. The processed command is then forwarded to the Verification layer, which is also implemented using the Phi-3.3B model. In this step, the framework performs validation to recognize mismatches between the user command and the current mission context, including unavailable targets, inconsistent task sequences, invalid navigation requests, or conflicts with environmental constraints. After successful verification, the validated command is transmitted to the Action Planning layer, implemented using Gemini 2.5 Flash, which performs hierarchical task decomposition and converts the high-level mission objective into a structured sequence of executable UAV actions, including takeoff, navigation, harvesting, and landing operations. 
An example output generated by the planner is shown below:

\begin{enumerate}
    \item takeoff uav1
    \item navigate uav1 to tree\_1
    \item harvest uav1 tree\_1
    \item navigate uav1 to tree\_2
    \item harvest uav1 tree\_2
\end{enumerate}
Following that, the created action plan is provided to the Dispatcher module, which translates the high-level task sequence into executable UAV control commands and coordinates the execution flow of navigation, perception, and harvesting activities during autonomous mission execution.

\subsubsection{Navigation module}

The navigation module is in charge of allowing the UAV to navigate autonomously around the orchard area. The resulting map incorporates spatial information about vacant space, inhabited regions, and surrounding impediments including tree trunks, branches, and canopy structures. Based on this environmental representation, the RRT* algorithm is utilized for global path planning. The planner incrementally explores the search space to generate an optimized collision-free trajectory from the current UAV position to the target tree location. During trajectory generation, collision checking is continuously performed to ensure that the generated path satisfies safety constraints and avoids environmental obstacles. Once a feasible trajectory is generated, the UAV autonomously follows the planned waypoints to reach the target tree : \[
T_i = (x_i, y_i, z_i),
\].


\subsubsection{Detection module}

After reaching the target tree location, the Dispatcher module continues the harvest action by assigning the UAV to traverse through three predefined bounding viewpoints surrounding the target tree. Each harvesting viewpoint is represented as:

\[
H_{\text{tree}_i} = \{H_1(x,y,z), H_2(x,y,z), H_3(x,y,z)\},
\]

where \(H_1\), \(H_2\), and \(H_3\) denote the three-dimensional bounding positions used for image acquisition around the \(i\)-th target tree. The UAV sequentially flies through these predefined viewpoints to capture RGB images from multiple perspectives, allowing improved visual coverage and reduced fruit occlusion under dense foliage conditions.

\begin{figure}[H]
\centering
\includegraphics[width=\linewidth]{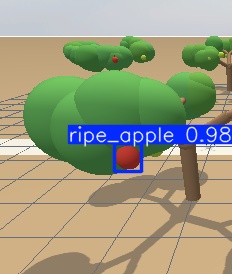}
\caption{The ripe fruit is detected by detection module}

\label{test:llm}
\end{figure}
The captured images are subsequently processed using the pretrained YOLOv8 object detection model for fruit recognition. Initially, the orchard image dataset is manually annotated with two classification labels: \textit{ripe} and \textit{unripe}. The labeled dataset is then utilized to train and fine-tune the YOLOv8 object detection model for fruit recognition. During the training process, the model learns discriminative visual features such as fruit color, texture, and shape characteristics to accurately distinguish harvestable ripe fruits from non-harvestable unripe fruits. After the training phase, the fine-tuned YOLOv8 model is deployed within the detection framework for real-time inference during UAV operation. During execution, the model simultaneously performs feature extraction, object location, and classification to generate bounding boxes and confidence scores for detected fruits based on RGB photos. Detection data from the three predetermined views are then pooled to increase fruit counting accuracy and decrease duplicate detections before being forwarded to the harvesting controller for autonomous harvesting.

\section{Environment Setup}

The proposed framework is developed and evaluated using a simulation-based UAV environment integrating Gazebo and PX4 SITL simulation \cite{px4_autopilot}. The UAV platform utilizes the \texttt{x500\_depth} model equipped with an RGB-D camera for simultaneous RGB image acquisition and depth perception during exploration and harvesting operations. PX4 is employed as the flight control stack, while QGroundControl is utilized for manual monitoring and map-building initialization during the exploration phase. Communication between Gazebo and ROS2 is established through the \texttt{ros\_gz\_bridge} interface, enabling real-time sensor data transmission and UAV state synchronization across the simulation framework.

For environment representation and visualization, the system integrates OctoMap for probabilistic voxel-based 3D mapping and RViz2 for real-time visualization of occupancy maps, navigation trajectories, and perception outputs. Voice commands are processed using the Whisper speech recognition model, while the multi-layer LLM framework consists of Phi-3.3B and Gemini-2.5-Flash models for edge-case analysis, command verification, and hierarchical task planning. The fruit detection module utilizes the YOLOv8 object detection framework fine-tuned for ripe and unripe fruit classification. Autonomous UAV control and trajectory execution are performed using MAVSDK communication between the mission planner and the PX4 flight controller.

The finished framework is run on a workstation equipped with an AMD Ryzen 7 Pro 6850U processor and 32 GB RAM, providing sufficient computational resources for real-time mapping, LLM inference, navigation planning, and computer vision processing.

For evaluating, the author 

\section{Evaluation}

To evaluate the performance of the proposed UAV harvesting system, ten experimental trials are conducted. Several quantitative metrics are defined to measure the accuracy of voice command interpretation, the effectiveness of fruit detection for harvesting, and the success rate of autonomous navigation.

The first metric evaluates the accuracy rate of voice command interpretation. This metric measures whether the system correctly converts the human voice command into the intended text-based instruction and generates the corresponding task plan. It is calculated as:

\[
A_{\text{voice}} = \frac{N_{\text{correct}}}{N_{\text{total}}} \times 100\%
\]

where \(N_{\text{correct}}\) denotes the number of correctly interpreted voice commands, and \(N_{\text{total}}\) denotes the total number of voice commands across all trials.

The second metric evaluates the effectiveness of fruit harvesting by measuring the rate of detected ripe fruits per tree. This metric determines how many ripe fruits are successfully detected by the YOLOv8-based detection module compared with the total number of ripe fruits present on the target tree. It is defined as:

\[
R_{\text{fruit}} = \frac{F_{\text{detected}}}{F_{\text{total}}} \times 100\%
\]

where \(F_{\text{detected}}\) represents the number of ripe fruits detected by the system, and \(F_{\text{total}}\) represents the total number of ripe fruits in the target tree.

The third metric evaluates the rate of successful navigation to the target. A navigation trial is considered successful when the UAV reaches the assigned target tree without collision and within an acceptable distance threshold from the target coordinate. This metric is computed as:

\[
S_{\text{nav}} = \frac{N_{\text{success}}}{N_{\text{trial}}} \times 100\%
\]

where \(N_{\text{success}}\) denotes the number of successful navigation trials, and \(N_{\text{trial}}\) denotes the total number of navigation trials.

\begin{figure}[H]
\centering
\includegraphics[width=\linewidth]{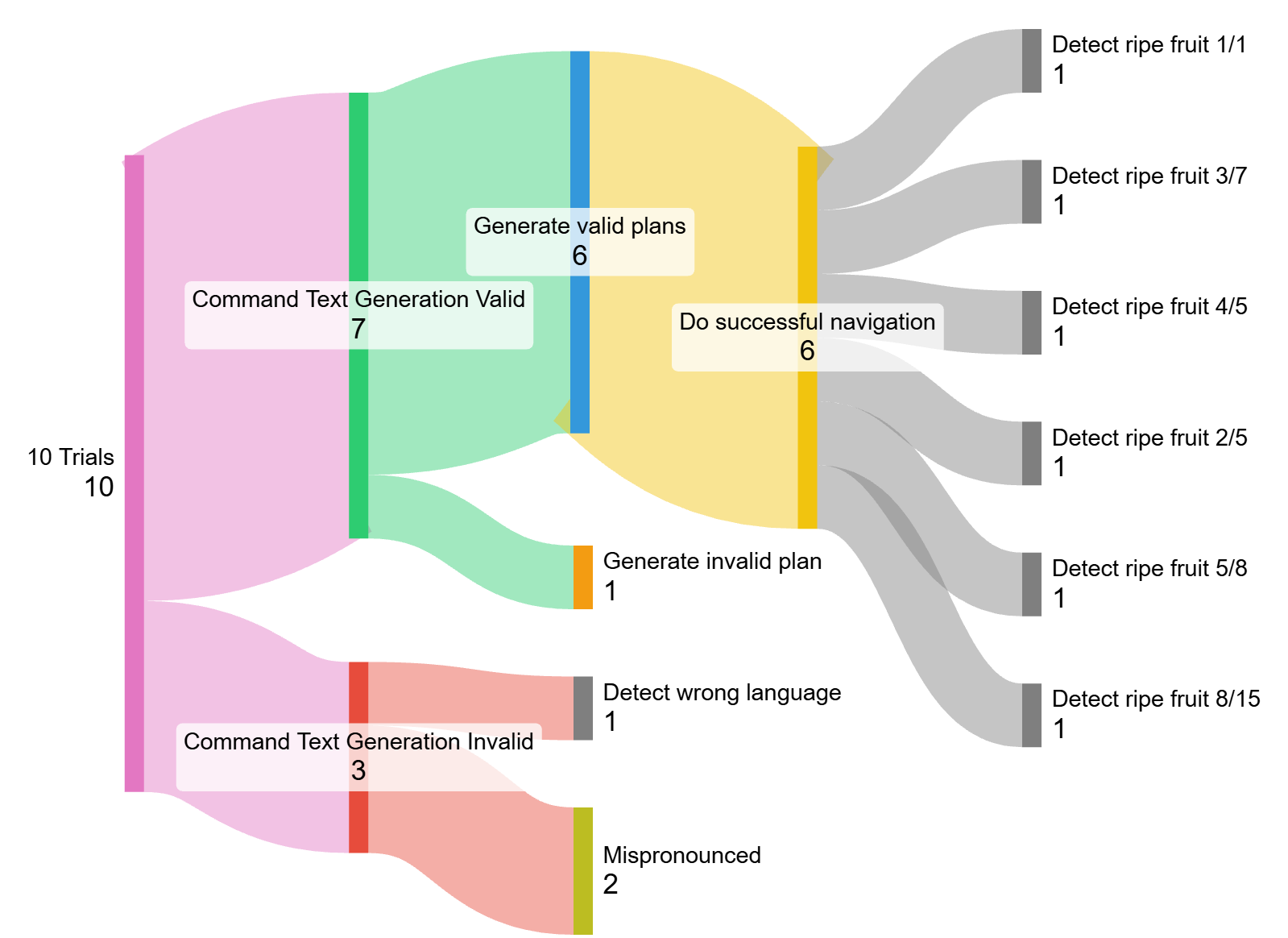}
\caption{Overall evaluation overview of the proposed framework across 10 experimental trials.}
\label{fig:evaluation_overview}
\end{figure}

In overview, Figure \ref{fig:evaluation_overview} presents the overall execution flow and performance distribution of the system across all trials. Among the ten trials, seven commands were successfully converted into valid text-based instructions by speech recognition module, while three commands failed due to incorrect language detection and pronunciation issues. Specifically, one failure occurred because the spoken instruction was understood in the incorrect language, while two failures were caused by imprecise pronunciation during voice input.

From the seven successfully interpreted commands, six generated valid task plans through the multi-layer LLM planning framework, while one generated an invalid execution plan during task verification. For the six valid mission plans, the navigation module successfully guided the UAV to the assigned target tree locations without collision. After reaching the target trees, the detection module executed ripe fruit detection u from three predefined viewpoints surrounding each tree.

The fruit detection results demonstrate varying detection performance depending on canopy density, fruit occlusion, and environmental visibility conditions. Across the six successful navigation trials, the system detected ripe fruits with detection rates of \(1/1\), \(3/7\), \(4/5\), \(2/5\), \(5/8\), and \(8/15\), respectively.

\begin{table}[H]
\centering
\caption{Performance metrics of the proposed UAV harvesting framework based on 10 experimental trials}
\label{tab:performance_metrics}
\begin{tabular}{|c|c|}
\hline
\textbf{Metric} & \textbf{Rate (\%)} \\
\hline
Voice Command Interpretation Accuracy& 70.0 \\
\hline
Valid Task Plan Generation Rate & 85.7 \\
\hline
Successful Navigation Rate& 100.0 \\
\hline
Average Ripe Fruit Detection Rate 
& 63.1 \\
\hline
\end{tabular}
\end{table}

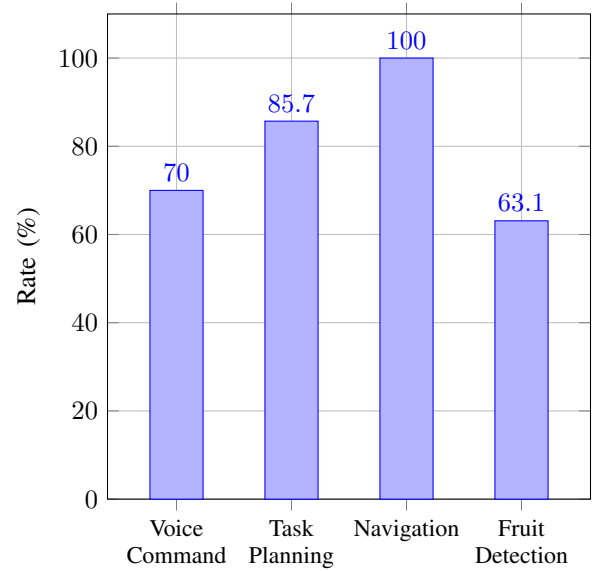
\begin{figure}[H]
\centering
\begin{tikzpicture}
\begin{axis}[
    ybar,
    bar width=20pt,
    width=0.9\linewidth,
    height=8cm,
    ymin=0,
    ymax=110,
    ylabel={Rate (\%)},
    symbolic x coords={
        Voice\\Command,
        Task\\Planning,
        Navigation,
        Fruit\\Detection
    },
    xtick=data,
    xticklabel style={
        align=center,
        font=\small
    },
    nodes near coords,
    nodes near coords align={vertical},
    enlarge x limits=0.2,
    grid=major,
]
\addplot coordinates {
    (Voice\\Command,70)
    (Task\\Planning,85.7)
    (Navigation,100)
    (Fruit\\Detection,63.1)
};
\end{axis}
\end{tikzpicture}
\caption{Comparison of evaluation metrics for the proposed UAV harvesting framework.}
\label{fig:metric_bar_chart}
\end{figure}

Among all evaluated components, the navigation module achieved the highest performance with a \(100\%\) successful navigation rate, indicating that the integration of OctoMap and RRT* provided reliable collision-free trajectory generation within the orchard environment. The task planning module also demonstrated strong performance with an \(85.7\%\) valid plan generation rate, showing that the multi-layer LLM architecture effectively handled semantic verification and hierarchical task decomposition. The voice command interpretation module achieved a \(70\%\) success rate,  caused during speech recognition. Compared with the other modules, the fruit detection module achieved the lowest performance with an average detection rate of \(63.1\%\). This reduction is mainly influenced by dense foliage, partial fruit occlusion, and varying illumination conditions within the orchard environment. 

Nonetheless, the overall results show that the proposed framework can perform end-to-end autonomous UAV harvesting operations by combining voice-guided interaction, intelligent task planning, autonomous navigation, and computer vision-based fruit detection into a single system.

\section{Limitation}
In overview, the framework demonstrates the possibility of voice-guided autonomous UAV harvesting, it also emphasizes the critical role of humans in the loop, allowing for a highly interactive operating process and bridging the gap between current autonomous technologies and agricultural applications. Nevertheless, several limitations of the suggested framework must be addressed.
\subsection{Fruit Detection and Deduplication}
The current detection module is unable to fully deduplicate ripe fruits detected from multiple viewpoints, which may lead to repeated detections of the same fruit instance during harvesting. In addition, the detection framework has difficulty recognizing fruits that are heavily occluded by branches and dense foliage, reducing the overall fruit detection accuracy under complex orchard conditions.
\subsection{Continuous Mission Execution and Replanning}
The system is currently limited in continuous mission execution, where long autonomous operations and dynamic mission replanning are not fully supported.

\subsection{Dialogue-Based Interaction}
Although, in task planning the system will ask the question if the command is not clear or does not specify any target trees, the system does not yet support dialogue-based interaction between the user and the system. This limitation will prevent the ability to perform perform interactive task clarification, contextual feedback, and real-time conversational mission updates.

\section{Future Work}
In conclusion, future system development should focus on prospective changes and enhancements that would strengthen the suggested framework's resilience, scalability, and practical usefulness. There are some suggestions the author wants to deep dive into:
\subsection{Robust Perception and Fruit Detection}

Future work will focus on improving the resilience and accuracy of the perception module. The current detection framework will be enhanced with object tracking algorithms to handle ripe fruit deduplication during harvesting missions. In addition, more advanced perception approaches will be investigated to better detect partially occluded fruits in dense canopy environments, where visual ambiguity and overlapping foliage remain significant challenges.

\subsection{Multi-Robot Collaboration}

The framework will also be extended to support multi-robot collaboration. Future developments will investigate coordinated task execution among multiple UAVs to improve harvesting efficiency, increase orchard coverage, and enable distributed task allocation for large-scale agricultural operations. For instance, a cooperative task-allocation system can be developed in which multiple drones autonomously distribute harvesting tasks based on fruit location, workload balancing, and mission priority to achieve faster and more intelligent harvesting operations. Another promising direction is a \textit{pick-and-catch} architecture, where an aerial robot is responsible for identifying and detaching ripe fruits while a ground robot positions itself to receive and transport the harvested fruits. Such collaboration can improve collection efficiency, and enhance the overall automation capability of orchard harvesting systems.

\subsection{Long-Term Autonomous Operation}

Another important aim is to allow continuous and long-term autonomous functioning. Future study will look into automatic replanning methods, adaptive task scheduling algorithms, and fault-tolerant mission management to allow persistent harvesting missions in dynamic environments.

\subsection{Human-Robot Interaction with LLM-Based Planning}

Finally, dialogue-based interaction capabilities will be integrated into the LLM planning framework. This enhancement aims to support conversational task clarification, contextual feedback exchange, and more natural human-robot interaction during mission execution. Incorporating interactive communication can further strengthen the role of humans in the loop while bridging the gap between current autonomous technologies and practical agricultural deployment.
\newpage
\bibliographystyle{IEEEtran}
\bibliography{references}
\end{document}